\documentclass[conference]{IEEEtran}
\IEEEoverridecommandlockouts
\usepackage[utf8]{inputenc}
\usepackage[T1]{fontenc}
\usepackage{amsmath,amssymb,amsfonts,amsthm}
\usepackage{mathtools}
\usepackage{algorithmic}
\usepackage{graphicx}
\usepackage[caption=false]{subfig}
\usepackage{textcomp}
\usepackage{hyperref}
\usepackage[noadjust]{cite}
\usepackage{xcolor}
\usepackage{amsmath}
\usepackage{array, multirow}
\usepackage{siunitx}
\usepackage{microtype}
\DeclareMathOperator*{\argmin}{argmin}
\DeclareMathOperator*{\minimize}{minimize}
\usepackage{environ}
\NewEnviron{optimization}[2][\minimize]{
  \setlength\arraycolsep{0.1em}
  \begin{align}
    #1\ &\begin{aligned}#2\end{aligned}\\
    \text{s.t.}\ &%
      \BODY
  \end{align}
}
\newtheorem{problem}{Problem}
\newtheorem{infproblem}{Informal Problem}
\newtheorem{solution}{Solution}

\allowdisplaybreaks

\def\BibTeX{{\rm B\kern-.05em{\sc i\kern-.025em b}\kern-.08em
    T\kern-.1667em\lower.7ex\hbox{E}\kern-.125emX}}

\begin{document}

\title{Physics Constrained Motion Prediction with Uncertainty Quantification
\thanks{This work was supported in part by NSF CCRI \#1925587 and DARPA \#FA8750-20-C-0542 (Systemic Generative Engineering). The views, opinions, and/or findings expressed are those of the author(s) and should not be interpreted as representing the official views or policies of the Department of Defense or the U.S. Government. This work is also supported in part by NSF GRFP Grant \#1845298}
}

\author{
        \IEEEauthorblockN{
            Renukanandan Tumu\IEEEauthorrefmark{1},
            Lars Lindemann\IEEEauthorrefmark{2},
            Truong Nghiem\IEEEauthorrefmark{3},
            Rahul Mangharam\IEEEauthorrefmark{1} 
        }
        \thanks{
            \IEEEauthorrefmark{1}
            \textit{University of Pennsylvania}
            Philadelphia, PA, USA 
            \{nandant, rahulm\}@seas.upenn.edu
        }
        \thanks{
            \IEEEauthorrefmark{2}
            \textit{University of Southern California},
            Los Angeles, CA, USA 
            llindema@usc.edu
        }
        \thanks{
            \IEEEauthorrefmark{3}
            \textit{Northern Arizona University}
            Flagstaff, AZ, USA
            Truong.Nghiem@nau.edu
        }
}

\maketitle

\begin{abstract}
Predicting the motion of dynamic agents is a critical task for guaranteeing the safety of autonomous systems. A particular challenge is that motion prediction algorithms should obey dynamics constraints and quantify prediction uncertainty as a measure of confidence. We present a physics-constrained approach for motion prediction which uses a surrogate dynamical model to ensure that predicted trajectories are dynamically feasible. We propose a two-step integration consisting of intent and trajectory prediction subject to dynamics constraints. We also construct prediction regions that quantify uncertainty and are tailored for autonomous driving by using conformal prediction, a popular statistical tool. Physics Constrained Motion Prediction achieves a 41\% better ADE, 56\% better FDE, and 19\% better IoU over a baseline in experiments using an autonomous racing dataset.
\end{abstract}


\section{Introduction}
A central problem in autonomous driving is in predicting the intents and future trajectories of dynamic agents. Planning a safe trajectory depends on accurate trajectory predictions of such agents. To predict the future motion of an agent, we must involve the agent's intent and its ability to realize that intent. To do so, we must understand the past motion of the agent and the surrounding context. This is complex due to the interactions between agents, their surroundings, and unknown objectives. To understand the ability to enact intent, we must factor in the ability of an agent to control itself under acceleration and steering limits. 

Further challenges in motion prediction are the potential for behaviors that are not present during training time and the presence of input noise, e.g.,  Zhang et. al. showed that motion prediction algorithms are susceptible to targeted noise attacks \cite{zhang_adversarial_2022}. There is hence a need for algorithms whose predictions are robust to noise and behavior shift.

% \todo[inline]{Is dynamic infeasibility a (critical) issue in motion prediction? If yes, cite and state it here, as it motivates our approach. - It has not been explicitly mentioned before as a critical issue, however, we are using dynamic feasibility as a proxy for what is considered a reasonable prediction. This method provides a guarantee that all of its predictions will be reasonable, which is more of a guarantee than is currently on offer. The value prop here is safety and trustworthiness related. - NT}

Physics constraints present a solution to this problem, restricting the output space of trajectories to only those that are physically possible. We use dynamics models, which describe the change in a system's state, to integrate these physics constraints. These constraints serve as a regularizer, encouraging better generalization. We propose Physics Constrained Motion Prediction (PCMP), an approach that uses a surrogate vehicle dynamic model and multi-step integration to ensure that all predicted motion is dynamically feasible under the given surrogate model. Specifically, we use neural networks, such as LSTMs, for the intent prediction step for which we have no conventional solution. We then use vehicle dynamics along with the predicted intent to predict trajectories. 

While this model decomposition gives us guarantees about dynamic feasibility, which we use as a proxy for reasonableness, it does not provide guarantees of accuracy. Therefore, we use conformal prediction to quantify prediction errors between PCMP and the unknown ground truth. Conformal Prediction provides the advantage of being data-driven, hence providing the ability to quantify uncertainty using statistical reasoning even if the surrogate model is not representative of the real system. We design two prediction regions tailored for uncertainty quantification in autonomous driving.

\begin{infproblem}
\label{prob:framing-one}
Given a series of past position observations ($O$), potentially containing location and orientation, and environmental context ($C$), potentially containing map information, predict the future positions ($F$) of an agent. We also wish to quantify the uncertainty of the predictions $\hat{F}$ such that they belong to the set $\mathcal{P}(\hat{F})$ with a desired probability.
\end{infproblem}

\begin{figure*}[t!]
  \centering
  \includegraphics[height=1.75in]{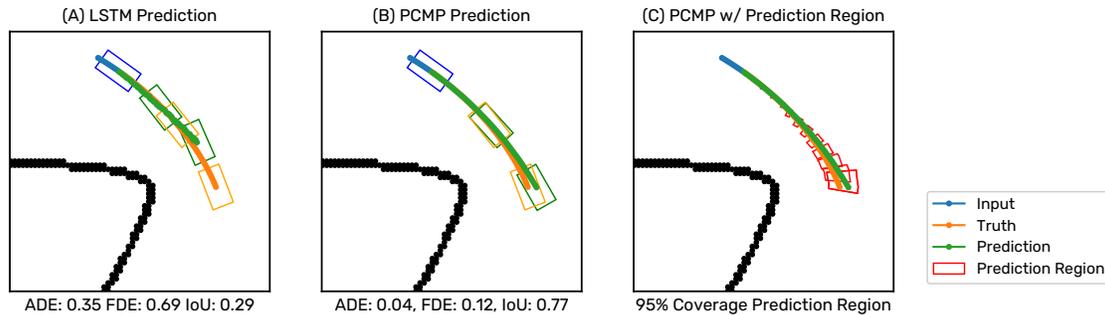}
  \vspace{-10pt}
  \caption{
  (A) shows an LSTM predictor with no physics information. We can see that the car is predicted to be sliding through the corner, and immediately reducing its speed. (B) shows a PCMP predictor with a bicycle model. The car is predicted to hold its speed through the corner and point in the direction of travel. (C) shows the PCMP predictor with conformal prediction regions. These regions are designed to contain the true trajectory 95\% of the time, showing us regions of the track to avoid. Bounding boxes are omitted in (C) for clarity. Prediction metrics are provided under each image.}
  \label{fig:ablation-diagram}
  \vspace{-15pt}
\end{figure*}

%\subsection{Contributions}
When solving the trajectory decoding task posed in Informal Problem \ref{prob:framing-one}, most approaches use a multi-layer perceptron (MLP), Long Short Term Memory network (LSTM)\cite{hochreiter_long_1997}, or other learned method to predict trajectories \cite[Table 1]{varadarajan_multipath_2022}. Our approach, PCMP, provides the following \textbf{contributions}:
\begin{enumerate}
    \item PCMP incorporates physical constraints by limiting control inputs to physical bounds and incorporating dynamics models into the trajectory generation. This \textbf{guarantees the dynamic feasibility} of predicted trajectories.
    \item PCMP provides \textbf{intent interpretability} through the prediction of control inputs that can be clearly decoded into intent; i.e, turning right and slowing down.
    \item PCMP is \textbf{robust to errors in parameter estimation} for its dynamics model.
    \item PCMP provides \textbf{informative prediction regions} via  two new scoring functions for conformal prediction.
    \item PCMP \textbf{performs better} than an LSTM baseline, with $41\%$ better ADE, $56\%$ better FDE, and $19\%$ better IoU.
\end{enumerate}

\subsection{Related Work}
%\subsubsection{Physics Constrained Neural Networks}
\emph{Physics-constrained neural networks} aim to constrain neural networks with physics biases \cite{karniadakis_physics-informed_2021}. Recent approaches use physics constraints and ODE solvers in the loop \cite{djeumou_neural_2022}, but aim to learn the dynamics of a system, constrained by known information about the system. Our approach differs from this approach by incorporating physical constraints through the use of activation functions, and by doing multi-step predictions instead of system identification.

%\subsubsection{Motion Prediction}
\emph{Motion Prediction} approaches can be broadly categorized into physics, pattern, and planning-based approaches, following the framework proposed in Karle et al. \cite{karle_scenario_2022}. Physics-based approaches use physical knowledge, and pattern-based approaches use learning and data to predict future motion. This work bridges the gap between pattern and physics-based approaches. This offers the dynamics guarantees of physics-based approaches, and the robustness to noise of pattern-based approaches, overcoming the disadvantages of each.

\paragraph{Physics based approaches}
The most simple physics-based approaches assume constant states like velocity and yaw rate, or some combination of these \cite{karle_scenario_2022}, like the Constant Turn Rate and Velocity (CTRV) model. Other methods use reachability analysis \cite{pek_using_2020} \cite{koschi_spot_2017}. These models, while they can be highly interpretable and offer guarantees of dynamic feasibility, can be brittle to noise and measurement error.

\paragraph{Pattern based approaches}
Pattern based-approaches to motion prediction focus contributions on context extraction from high-definition maps and positional information, which in our problem formulation, is the context $C$. Motion prediction approaches use LSTMs, graph neural networks, and transformers, to extract map information \cite{alahi_social_2016}, \cite{gilles_thomas_2022}, \cite{liang_learning_2020}, \cite{nayakanti_wayformer_2022}. These approaches are the current state of the art on modern motion prediction datasets \cite{chang_argoverse_2019}. They are strong in prediction in noisy environments and are capable of understanding intent given context, however the problem they seek to solve centers around useful context extraction from multi-modal data. 

When solving the trajectory decoding task, pattern-based approaches use time series learning approaches like LSTMs \cite{varadarajan_multipath_2022}. These approaches do not provide any guarantees of dynamic feasibility. Trajectron++ \cite{salzmann_trajectron_2021} attempts to combine the two approaches, by pairing a pattern-based encoding with a dynamics-based decoder.
Our method differs from Trajectron++ in the following ways. PCMP does not use an autoencoder or other mechanism to propagate gradients to the feature extractor without using the dynamics. PCMP instead propagates all gradients through the dynamics integration steps. Additionally, PCMP is able to handle the bicycle model, which is more specific than the unicycle model and challenging during backpropagation due to the $\tan$ function and division in the bicycle model. With some of the same tricks we propose, however, Trajectron++ could potentially be extended to the Bicycle model.  PCMP offers a curriculum training approach for models that incorporate dynamics integration for more efficient training, and, finally, PCMP quantifies uncertainty using Conformal Prediction.

%\subsubsection{Conformal Prediction}
\emph{Conformal prediction} is a statistical tool for uncertainty quantification that has recently been explored in machine learning applications \cite{angelopoulos_gentle_2022}. An independent and identically distributed (i.i.d.) validation dataset is used to construct prediction regions in which the unknown true value will be contained with a desired probability of $1-\delta$. These regions are constructed by calculating the $1-\delta$ quantile of a scoring function, e.g., the displacement between the prediction and the ground truth, over the empirical distribution of the validation dataset. Of particular interest to us are conformal prediction frameworks for motion prediction \cite{stankeviciute_conformal_2021} and planning under motion prediction uncertainty \cite{lindemann_safe_2022,dixit_adaptive_2022}.  We explore scoring functions suitable for driving applications and use a refined technique called Conformalized Quantile Regression (CQR) \cite{romano_conformalized_2019}. To the best of the authors' knowledge, this is the first work that creates vehicle and map-based prediction regions.

\section{Methodology}
\begin{figure*}[t]
  \centering
  \includegraphics[height=2in]{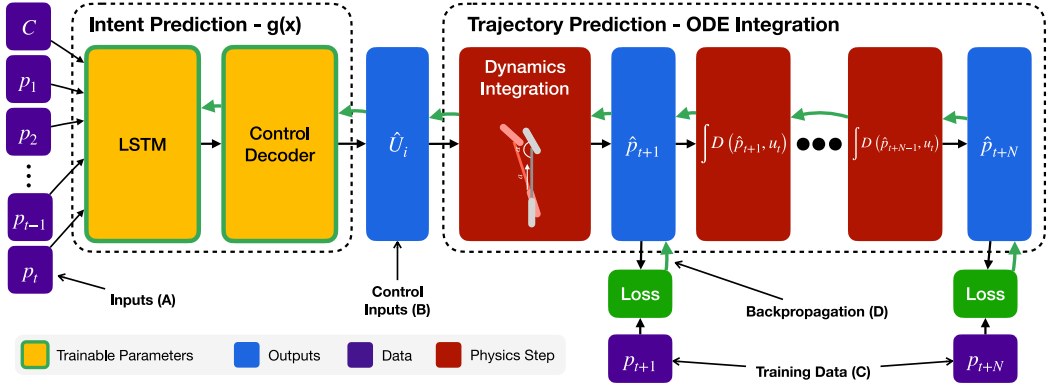}
  \vspace{-5pt}
  \caption{This figure shows the PCMP prediction and training algorithm. The inputs (A), are fed into the LSTM, with the context vector $C$ concatenated to each position. The hidden state of the LSTM is processed through an MLP and scaled based on control input constraints to generate control inputs (B). These control inputs are then passed into the dynamics integration step, which solves the ODEs of the dynamics equations with our control inputs and the last known position of the vehicle for the prediction horizon. At training time, these predicted positions are then used to calculate a loss with the training data (C). This loss is backpropagated (D) through the Trajectory Prediction steps and used to train the Intent Prediction networks.}
  \label{fig:pimp-overview}
  \vspace{-15pt}
\end{figure*}

PCMP consists of three parts. First, we use pattern-based methods to understand the intent of the agent, in the form of  control inputs. We impose limits on the control inputs, which correspond to physical constraints. We call this step the intent prediction step. Next, we use our dynamics model and the predicted control inputs to predict the trajectory the vehicle will take. We formalize this below. Afterward, we use Conformal Prediction to build prediction regions in a particular deployment domain. These prediction regions provide probabilistic guarantees of where the true trajectory will lie, helping us understand the uncertainty in our predictions.

\begin{problem}
We are given a set of past observed positions of an agent $O_i = \left[p_{i,t-l}, \dots, p_{i,t}\right]$ at some time $t$, a time-invariant environmental context variable $C_i$, and the future positions of the agent $F_i = \left[p_{i,t+1}, \dots, p_{i,t+n}\right]$. Here, $l$ is the observation horizon and $n$ is the prediction horizon.

We collect these in a training dataset $\mathcal{D}_{\textrm{train}} = \{(O_0, C_0, F_0), \ldots, (O_{N_t}, C_{N_t}, F_{N_t})\}$ and validation and test datasets $\mathcal{D}_{\textrm{val}}$ and  $\mathcal{D}_{\textrm{test}}$, which are similarly composed. We would now like to find an optimal motion prediction model $f(O_i, C_i)$ such that:
\begin{equation}
    f = \argmin_f \sum_{i=0}^{|\mathcal{D}_{\textrm{test}}|} \left\|F_i - f(O_i, C_i)\right\|_1
\end{equation}
Further, we would like to create a prediction region $\mathcal{P}_{val}$ such that the true future trajectory is contained in it with a probability of at least $1-\delta$
\begin{equation}
    \mathbf{P}\left(F_{test} \in \mathcal{P}_{\textrm{val}}\right) \geq 1 -\delta
    \label{eq:conformity-guarantee}    
\end{equation}
\end{problem}
\begin{solution}
We use surrogate vehicle dynamics $D(p,u) = \dot{p}$, to constrain agent motion, and sampling time $ts$, which defines the time between two subsequent observations. We set $f$ to be equal to the output $\hat{F} = [\hat{p}_{i,t+1}, \ldots, \hat{p}_{i,t+n}]$, which is found by solving the optimization problem described below, with a weight vector $\lambda$.

\begin{optimization}[\minimize_g]{\sum_{i=0}^{|\mathcal{D}_{\textrm{train}}|} \mathcal{L}\left( F_i , \begin{bmatrix} \hat{p}_{i,t+1} \\ \vdots \\ \hat{p}_{i,t+n} \end{bmatrix} \right)}
\hat{p}_{i,t} = p_{i,t}  \label{eq:int-step1} \\
&\hat{p}_{i,j+1} = \hat{p}_{i,j} + \int_0^{ts} D(\hat{p}_{i, j}, \hat{u}_{i,j})\,dt \nonumber \\ 
&\qquad \quad t<j<t+n-1\label{eq:int-step2}\\
& \begin{bmatrix} \hat{u}_{t} & \ldots & \hat{u}_{t+n-1} \end{bmatrix}^\top = U_i = g(O_i, C_i) \label{eq:g-function}\\
&\mathcal{L}(F_i, \hat{F}_i) = \frac{1}{n}\sum_{i=t+1}^{t+n} \left|\lambda\cdot\left(F_i - \hat{F}_i\right)\right| \label{eq:loss-func}
\end{optimization}

The uncertainty quantification problem, on the other hand, is solved using Conformalized Quantile Regression (CQR) \cite{romano_conformalized_2019} as detailed later, which will provide us the confidence measure described in \eqref{eq:conformity-guarantee}. We assume that $\mathcal{D}_{\textrm{val}}$ and  $\mathcal{D}_{\textrm{test}}$ are drawn from the same distribution.
\end{solution}

Equations \eqref{eq:int-step1} sets our initial condition, and \eqref{eq:int-step2} integrates the dynamics. The prediction of control inputs is shown in \eqref{eq:g-function}. Note that instead of optimizing for $f$, we only optimize for the intent prediction $g$. Because we are only given $O$ and $F$, we must complete the integration of the dynamics, using the last known vehicle state as given in $O$, and solve our minimization problem by backpropagating our losses through the dynamics. This has advantages; first, it does not require the solution of a control problem to obtain $U$, and second, it allows us to use observed data from a deployed system directly, without the need for pre-processing.

\subsection{Intent and Trajectory Prediction \& Training}

\emph{Intent prediction} is represented by the function $g$, described in Equation \eqref{eq:g-function}, and depicted in Figure \ref{fig:pimp-overview}. Classes of function approximators that can be used to approximate $g$ include LSTMs \cite{hochreiter_long_1997}, Transformers \cite{vaswani_attention_2017}, Graph Neural Networks, or other neural network architectures. In our experiments, we use an LSTM to predict intent. Positive acceleration means the vehicle is predicted to speed up, and negative means braking intent. Negative steering indicates a left turn, positive indicates a right turn.

Because our outputs are control inputs, which have real physical analogs, we must constrain our final output layer to match the physical limits of the agent. The $\tanh$ function scales values from a domain of $\mathbb{R}$ to a range of $(-1,1)$. Given an activation $a$, we define our activation function $\phi_{\omega}(a) = \omega \tanh(a)$. This aids in training, when values like acceleration and steering angle with different ranges must be predicted by the same network. We can also use this to impose constraints on the magnitude of the control outputs, for example: limiting the range of the steering angle. These control inputs can be read and interpreted to understand vehicle intent.

For \emph{trajectory prediction}, we select surrogate dynamics $D(p, u) = \dot{p}$ to represent the motion of vehicles in our scenario. We use this dynamics function by integrating it to produce our output, as shown in the Trajectory Prediction section of Figure \ref{fig:pimp-overview}. Because we are given $x_t$, an iterative update rule, described in \eqref{eq:int-step1} and \eqref{eq:int-step2}, is used to propagate the system forward. The integration is done numerically, using either the Runge-Kutta method or the Euler method. We found that in cases where the timesteps are further apart than $0.1$ seconds, RK4 integration presented an advantage. We can use any dynamics model here, so long as attention is paid to areas of the state and input space where numerical instabilities emerge. The $\tan$ and division operators are some examples of sources of numerical instability.

\label{sec:dyn-feas-guarantee}
Due to our use of a dynamics model, all predictions from this method are dynamically feasible, regardless of the control input. We define a dynamically feasible state transition as a transition between $p_t \to p_{t+1}$ with timestep $ts$, according to a dynamics model $D(p,u) = \dot{p}$ which is dynamically feasible if $\exists u: p_t + \int_0^{ts} D(p_t, u)\,dt = p_{t+1}$. A trajectory $p_0, p_1, \ldots, p_n$ is said to be dynamically feasible according to the model $D$ and timestep $ts$ if each of the transitions $p_i \to p_{i+1}$ are dynamically feasible. This is true by construction in our method, as the update rule \eqref{eq:int-step2} implements a state transition.

For \emph{training of PCMP}, we evaluate $\hat{F}_i$ using  Equations \eqref{eq:int-step1}-\eqref{eq:g-function} for each item in our dataset and optimize using stochastic gradient descent. We use the L1 loss function on the difference between $\hat{F}_i$ and $F_i$, which is multiplied by the weight vector $\lambda$. The loss function is given in \eqref{eq:loss-func}. The L1 loss was experimentally found to be more effective for training than the L2 loss.

%\subsubsection{Increasing Horizon Curriculum}
We use an increasing horizon curriculum to generate better predictions. This is accomplished by calculating the loss using the modified loss function in Equation \eqref{eq:curriculum-loss}, and increasing $h$ periodically from $0\leq h \leq n$ with $n$ as the prediction horizon.
\begin{equation}
    \mathcal{L}_h(F_i, \hat{F}_i) = \frac{1}{h}\sum_{i=t+1}^{t+h} \left|\lambda\cdot\left(F_i - \hat{F}_i\right)\right|
    \label{eq:curriculum-loss}
\end{equation}

\subsection{Uncertainty Quantification with Conformal Prediction}
\label{sec:cp-regions}
\begin{figure}
    \centering
    \includegraphics[width=0.75\linewidth]{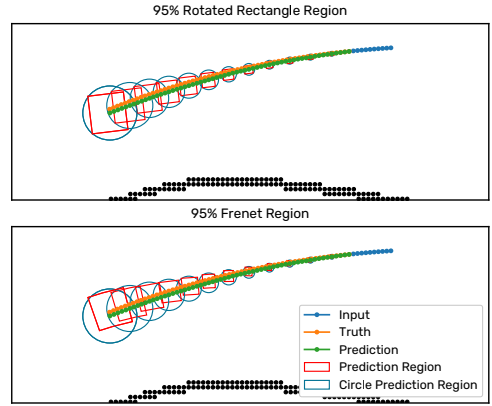}
    \vspace{-10pt}
    \caption{The proposed conformal prediction regions. Top: the orientation of the rectangles rotates with the last known position of the vehicle, which is the last blue point in the figure. Bottom: the prediction regions in the Frenet Region diagram show boxes that curve along with the shape of the track.}
    \label{fig:cp-regions}
    \vspace{-10pt}
\end{figure}

Conformal prediction provides a simple method for quantifying prediction uncertainty \cite{angelopoulos_gentle_2022}. It is based on a scoring function that evaluates how well a given prediction approximates the true but unknown value. Other works that quantify uncertainty use circle-shaped prediction regions that are not necessarily optimal for autonomous vehicles. We propose two scoring functions that are tailored for applications in driving with the Rotated Rectangle Region and the Frenet Region as our scoring functions. Examples of these output shapes are shown in Figure \ref{fig:cp-regions}, along with the baseline circle prediction region. All of the regions are generated with the same $1-\delta$ of $95\%$.

\subsubsection{Rotated Rectangle Region}
The Rotated Rectangle Region is a Conformal Prediction Region where the region takes the shape of a rectangle that has been rotated to match the orientation of the last pose of the vehicle. This region preserves the vehicle's direction of travel, revealing directional biases in the predictor where an unrotated rectangle would not. This method works even when a map of the environment is not available.

To calculate the rotated rectangle region, we change the coordinate frame of the prediction to be the last input pose of the vehicle, $p_t$. We also normalize the rotation of the vehicle at that point, so the last known pose of the vehicle is identical across each prediction. We denote the transformed position vector $\prescript{L}{}{p_{i,t}} = [\prescript{L}{}{x_{i,t}}, \prescript{L}{}{y_{i,t}}, \prescript{L}{}{\theta_{i,t}}]$, which designates that $x, y, \theta$ are represented in the frame of the last known position of the vehicle. For each predicted point, we calculate the signed error in the X and Y direction, giving us the scoring function 
$$
s(\prescript{L}{}{\hat{p}_{i,t}}, \prescript{L}{}{p_{i,t}}) = \begin{bmatrix}
    \prescript{L}{}{x_{i,t}}-\prescript{L}{}{\hat{x}_{i,t}} \\
    \prescript{L}{}{y_{i,t}}-\prescript{L}{}{\hat{y}_{i,t}}
\end{bmatrix}
$$

\subsubsection{Frenet Region}
In cases where we do know the map or track, we can use the Frenet Region, which calculates errors in the Frenet coordinate frame. The Frenet frame is defined by a progress variable $s$, and a displacement $d$. The progress takes values on some interval usually $[0,1]$, and specifies the closest location on the track centerline to the point. $d$ is a signed displacement that gives us the distance from the point to the closest centerline point. These regions, by virtue of the Frenet frame, will follow the curves of the track. To create the Frenet Region, the prediction and target coordinates are converted to the Frenet coordinate frame. We denote the Frenet coordinates by $\prescript{F}{}{p_{i,t}} = [\prescript{F}{}{s_{i,t}}, \prescript{F}{}{d_{i,t}}]$, and the scoring function:
$$
s(\prescript{F}{}{\hat{p}_{i,t}}, \prescript{F}{}{p_{i,t}}) = \begin{bmatrix}
    \prescript{F}{}{s_{i,t}} - \prescript{F}{}{\hat{s}_{i,t}} \\
    \prescript{F}{}{d_{i,t}} - \prescript{F}{}{\hat{d}_{i,t}}
\end{bmatrix}
$$

%\subsubsection{Distributional Challenges in Conformal Prediction}\todo[]{Workshop Section Title}
\subsubsection{Conformal Prediction}
At its core, conformal prediction uses a validation dataset that consists of i.i.d. data points. Recall that the elements of this dataset
are past observations $p_{\textrm{pre}}^t=[p_{t-l}, \ldots, p_t]$ and future states $p_{\textrm{post}}^t=[p_{t+1}, \ldots, p_{t+n}]$. These data points can be thought of as drawn from a distribution $\mathcal{D}_t$, i.e., $[p_{\textrm{pre}}^t, p_{\textrm{post}}^t] \sim \mathcal{D}_t$. It is  easy to see that the elements of $\mathcal{D}_\text{cal}$ are not i.i.d. and hence violate the i.i.d. assumption made in conformal prediction. The reasons for this are twofold and we will alleviate this issue as follows.

First, there is a natural coupling between neighboring data points such as $[p_{\textrm{pre}}^t, p_{\textrm{post}}^t]$ and $[p_{\textrm{pre}}^{t+1}, p_{\textrm{post}}^{t+1}]$ that results in dependence between the distributions $\mathcal{D}_t$ and $\mathcal{D}_{t+1}$. To remedy this dependence in practice, we create a validation dataset with elements that do not share the same positions so that every position in a vehicle trace was only used once. While this does not fully eliminate the issue of dependence, we found that it works  well in practice. Second, the distributions $\mathcal{D}_t$ and $\mathcal{D}_{t'}$ may  be different because they were observed in different parts of the track. Ultimately, we are interested in the independence of the scoring function $s(\hat{p}_{i,t}, p_{i,t})$. In our validation dataset, we have created a stratified sample on the race lines and the controllers, which should ensure that the data points in the validation and test datasets are not strongly correlated. 

Having discussed these challenges, we can now apply CQR. CQR is a form of Conformal Prediction that provides two-sided prediction intervals. We will only present the algorithm in the interest of space, and refer the interested reader to \cite{angelopoulos_gentle_2022} and \cite{romano_conformalized_2019} for more information. First, we calculate the scoring function $s(\hat{p}_{i,t}, p_{i,t})$ for each timestep and item in the training dataset. We then calculate the lower and upper quantile bounds $[q_{\textrm{low},t}, q_{\textrm{high},t}]$ corresponding to the $\bar{\delta}/2$ and $1-(\bar{\delta}/2)$ quantiles over the distribution of $s(\hat{p}_{i,t}, p_{i,t})$. This gives us a baseline for our prediction region for which we define the non-conformity score $R$ for each item in the validation dataset as
\[
    R_{i,t} = \max\left\{q_{\textrm{low},t} - s(\hat{p}_{i,t}, p_{i,t}), s(\hat{p}_{i,t}, p_{i,t}) - q_{\textrm{high},t}\right\}
\]
This non-conformity score can be thought of as an error metric for how conservative or liberal our prediction regions are. We then pick the $(1-\bar{\delta})(1+ 1/|\mathcal{D}_{\textrm{val}}|)$ quantile of the non-conformity scores, and call this $E_{1-\bar{\delta},t}$. Under the assumption that the elements of the validation data are i.i.d., our single-step prediction region with a $(1-\delta)$ coverage guarantee is $\mathcal{P}_{\textrm{val},t} = [q_{\textrm{low},t} - E_{1-\bar{\delta},t}, q_{\textrm{high},t} + E_{1-\bar{\delta},t}]$ were we set $\bar{\delta} = \delta/2$ to union bound over the dimensions in our nonconformity score. Our choice of $\bar{\delta}$ is particularly important when we want coverage guarantees for multi-step predictions. If we want to be assured that $\mathbf{P}\left(p_{i,t} \in \mathcal{P}_{t,\textrm{val}} \forall t )\right) \geq 1 -\delta$, we can achieve this by setting $\bar{\delta} = \delta/n$ \cite{lindemann_safe_2022}.

\section{Results And Discussion}
We ran experiments to evaluate the performance of PCMP. In Section \ref{sec:performance}, we show that the performance of PCMP is better than the LSTM and CTRV baselines in all measured metrics. In Section \ref{sec:ood-driving}, we show that the PCMP predictor provides better performance than the baselines on behaviour it has not seen before. Section \ref{sec:model-error} shows that underestimating the model parameter can provide performance improvements, meaning that we need not be perfect in estimating the model parameters of vehicles we hope to predict for.

\subsection{Performance Measures}
\label{sec:performance-measures}
Average and Final Displacement Errors (ADE and FDE), were used to capture the correctness of the trajectories. ADE is computed by taking the average of the displacement between predicted and true points, and FDE is computed by examining the displacement between the final predicted and true points. These metrics do not account for predicted heading error, however, they are the standard metrics used in the field, so we adopt them.

We additionally use Intersection over Union (IoU), a metric commonly used in image segmentation and object detection. This metric is calculated by overlaying the two bounding boxes for the vehicle, and calculating the ratio of the intersecting area of the prediction and ground truth to the union of the prediction and true values.

\subsection{Dataset}
\label{sec:dataset}
\begin{figure}[t]
   \centering
   \subfloat[Race lines used to generate the dataset, on the track Spielberg using the F1Tenth Gym Simulator, in corners 4, 5, and 6.\label{fig:dataset-racelines}]{\includegraphics[width=0.45\linewidth]{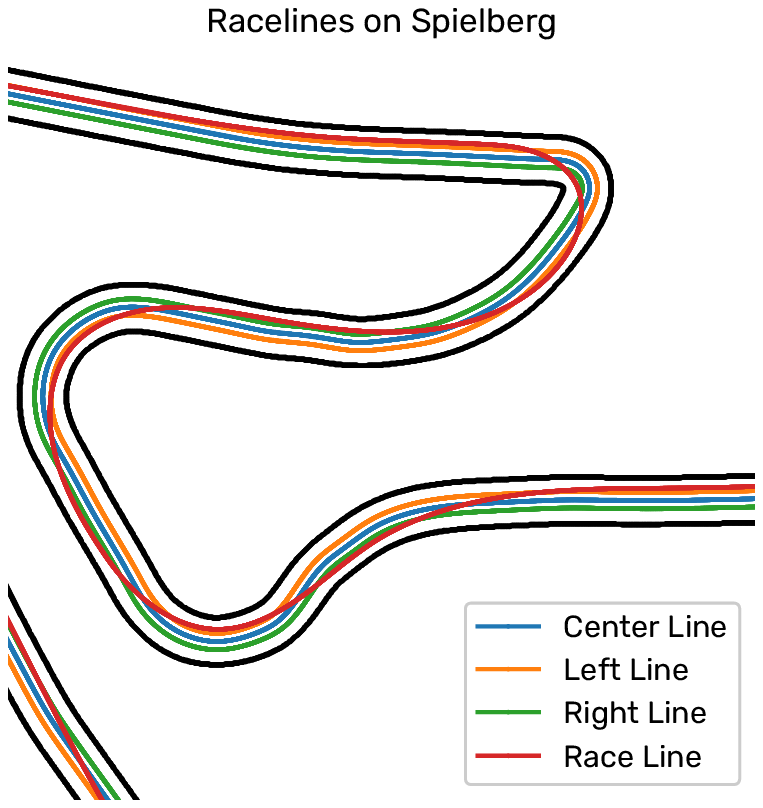}}\quad
   \subfloat[Plot (A) shows an LSTM prediction in OOD dataset, with a spinning behavior. (B) shows the same scenario with a PCMP prediction.\label{fig:lstm-ood-error}]{\includegraphics[width=0.45\linewidth]{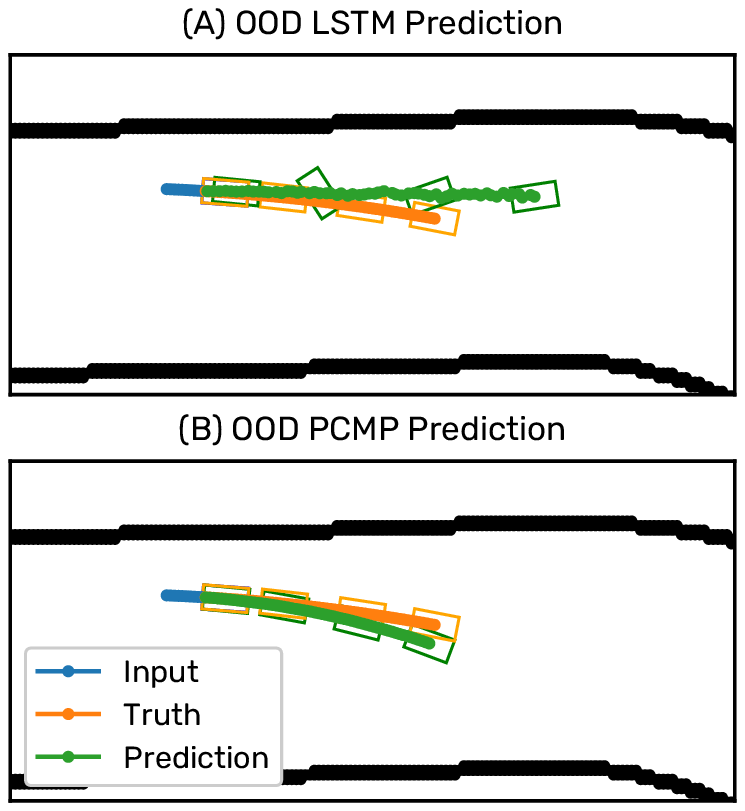}}
   % \begin{subfigure}[c]{0.45\linewidth}
   %      \centering
   %      \includegraphics[height=\textwidth]{images/track_image.eps}
   %      \caption{Race lines used to generate the dataset, on the track Spielberg using the F1Tenth Gym Simulator, in corners 4, 5, and 6.}
   %     \label{fig:dataset-racelines}
   % \end{subfigure}
   % \begin{subfigure}[c]{0.45\linewidth}
   %     \centering
   %     \includegraphics[height=\textwidth]{images/robustness-error.eps}
   %     \caption{Figure showing LSTM prediction for OOD dataset, with spinning behavior}
   %     \label{fig:lstm-ood-error}
   % \end{subfigure}
   \vspace{-15pt}
\end{figure}

Our dataset was created in the F1Tenth Gym \cite{okelly_f1tenth_2020}, a simulator for autonomous racecars, which was extended with RK4 integration for this work. We created the training set by observing racecars following a number of predefined trajectories on the track Spielberg. These trajectories fell into two categories, centerline offsets, and an optimized race line, pictured in Figure \ref{fig:dataset-racelines}. There are three centerline offsets; the centerline itself, and a left and right offset. The optimal race line was generated using the TUM race line optimization toolkit \cite{heilmeier_minimum_2020}. This optimized race line exceeds the bounds of all three centerline offsets, and we consider it to be out of the distribution of centerline offsets. These controllers were followed by two different controllers, the Stanley Controller \cite{hoffmann_autonomous_2007}, and a Pure Pursuit controller. Input samples are selected by taking non-overlapping 10 sample regions of the trace as input. The ensuing 60 samples are used as the output.

The training and testing splits are stratified by trajectory and controller, so all race line-controller pairs are proportionally represented in the training, validation, and testing sets. $O_i$ contains 0.1 seconds of position and velocity: $(x,y,\theta, v)$ sampled at 100 Hz, and $F_i$ contains 0.6 seconds of position and velocity data as the output. $C_i$ contains the forward curvature of the track at the last point in $O_i$. To approximate measurement noise, a noise vector $\epsilon \sim \mathcal{N}(0, 0.01)$ is added to the position and velocity.

\begin{table}[h]
   \vspace{-10pt}
   \caption{Number of samples in autonomous racing dataset}
   \label{tab:dataset-size}
   \centering
   \tiny
   \begin{tabular}{| c | c | c | c | c | c |}
       \hline
       \multicolumn{1}{|c|}{\multirow{2}{*}{Race Line}} & \multicolumn{1}{c|}{\multirow{2}{*}{Controller}} & \multicolumn{1}{c|}{\multirow{2}{*}{Speeds}} & \multicolumn{3}{c|}{Sample Counts}\\
       \cline{4-6}
       \multicolumn{1}{|c|}{}&\multicolumn{1}{c|}{}&\multicolumn{1}{c|}{}&\multicolumn{1}{c|}{Training}&\multicolumn{1}{c|}{Validation}&\multicolumn{1}{c|}{Test}\\
       \hline
       \multirow{6}{*}{Center}  & \multicolumn{1}{c|}{\multirow{3}{*}{Pure Pursuit}}    & 0.75 & 1684 &  210 & 211 \\ \cline{3-6} 
                                &                                                       & 0.85 & 1486 &  186 & 186 \\ \cline{3-6}
                                &                                                       & 1.00 & 1264 &  158 & 158 \\ \cline{2-6} 
                                & \multicolumn{1}{c|}{\multirow{3}{*}{Stanley}}         & 0.75 & 1689 &  211 & 211 \\ \cline{3-6} 
                                &                                                       & 0.85 & 1492 &  186 & 186 \\ \cline{3-6}
                                &                                                       & 1.00 & 1268 &  159 & 159 \\ \hline
       \multirow{6}{*}{Left}    & \multicolumn{1}{c|}{\multirow{3}{*}{Pure Pursuit}}    & 0.75 & 1719 &  215 & 215 \\ \cline{3-6} 
                                &                                                       & 0.85 & 1517 &  190 & 190 \\ \cline{3-6}
                                &                                                       & 1.00 & 1291 &  161 & 161 \\ \cline{2-6} 
                                & \multicolumn{1}{c|}{\multirow{3}{*}{Stanley}}         & 0.75 & 1724 &  215 & 215 \\ \cline{3-6} 
                                &                                                       & 0.85 & 1521 &  190 & 190 \\ \cline{3-6}
                                &                                                       & 1.00 & 1294 &  162 & 162 \\ \hline
        \multirow{6}{*}{Right}  & \multicolumn{1}{c|}{\multirow{3}{*}{Pure Pursuit}}    & 0.75 & 1644 &  205 & 206 \\ \cline{3-6} 
                                &                                                       & 0.85 & 1452 &  181 & 182 \\ \cline{3-6}
                                &                                                       & 1.00 & 1236 &  155 & 154 \\ \cline{2-6} 
                                & \multicolumn{1}{c|}{\multirow{3}{*}{Stanley}}         & 0.75 & 1654 &  207 & 207 \\ \cline{3-6} 
                                &                                                       & 0.85 & 1430 &  186 & 183 \\ \cline{3-6}
                                &                                                       & 1.00 & 1244 &  156 & 155 \\ \hline
        \multirow{6}{*}{Race}   & \multicolumn{1}{c|}{\multirow{3}{*}{Pure Pursuit}}    & 0.75 & 1528 &  191 & 191 \\ \cline{3-6} 
                                &                                                       & 0.85 & 1348 &  169 & 169 \\ \cline{3-6}
                                &                                                       & 1.00 & 1145 &  143 & 143 \\ \cline{2-6} 
                                & \multicolumn{1}{c|}{\multirow{3}{*}{Stanley}}         & 0.75 & 1530 &  191 & 191 \\ \cline{3-6} 
                                &                                                       & 0.85 & 1349 &  169 & 169 \\ \cline{3-6}
                                &                                                       & 1.00 & 1147 &  143 & 143 \\ \hline \hline
        Total & & & 34686 & 4336 & 4336 \\
        \hline
   \end{tabular}
   %performance-comparison/(LSTM-CURVATURE-8|PIMP-30.*-8-.*-CURR-1)
   \vspace{-10pt}
\end{table}

\subsection{Experimental Parameter Settings}
\label{sec:parameters}
For our baseline, we create an LSTM-based predictor, with a hidden state size of $16$ and an MLP decoder to predict the states. In terms of our problem statement, we approximate $f(O, C)$ with an LSTM. This is a fair baseline, as the map information and positional information is already extracted. The exact same framework is used for the physics-constrained model, where the same LSTM is used to approximate $g(O, C)$. Both models perform well on the autonomous racing dataset.

The model we used was the kinematic bicycle model, with the control point defined at the center of the line connecting the midpoints of the front and rear axles. We define $p = [x, y, \theta, v]^\top$, where $x,y$ are position, $\theta$ and $v$ are heading and velocity. $u= [\delta, a]^\top$, where $\delta$ is steering angle and $a$ is acceleration. The model has a single parameter, the wheelbase $L$, which is set to the true value, $0.3302$.
\begin{equation}
    D(p, u) = \dot{p} = \begin{bmatrix}
        \left(v+\frac{L}{2}\right)\cos(\theta)\\[0.5ex]
        \left(v+\frac{L}{2}\right)\sin(\theta)\\[0.5ex]
        \frac{v\tan(\delta)}{L}\\[0.5ex]
        a
    \end{bmatrix}
\end{equation}

We scaled the outputs of the Intent Prediction network by scaling the values of the acceleration and steering angle in the ranges $(-20, 20)$, $[- 7\pi /16,- 7\pi /16]$ respectively. We bounded the steering angle lower than $\pm \pi/2$ due to the numerical instability of the tangent function about that value.

We set $\lambda = [1, 1, 4, 0]$ in our loss function, multiplying the heading error by $4$, and ignoring the velocity error. For the curriculum approaches, we incremented $h$ by one every two epochs in the loss function defined in Equation \ref{eq:curriculum-loss}. The models were trained for $350$ epochs.

\subsection{Performance}
\label{sec:performance}
To start, we evaluate the performance of our algorithm vs the LSTM. All approaches were trained for 1500 epochs on all race lines and evaluated on all race lines. The ADE and FDE on the test set are shown in Table \ref{tab:performance-results}. PCMP achieves $41\%$ better ADE, $56\%$ better FDE, and a $19\%$ increase in IoU.

\begin{figure}[h]
   \centering
   \includegraphics[width=0.75\linewidth]{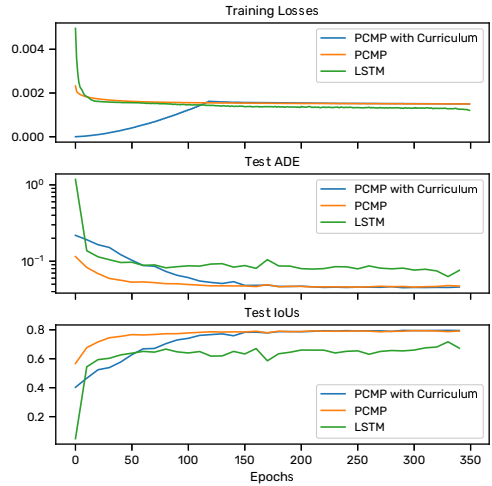}
   \vspace{-10pt}
   \caption{Plot showing that despite the LSTM approach achieving a lower training loss, the PCMP methods achieve lower displacement error and IoU error, which takes heading error into account, on the test set. In the plot showing training losses, the blue line representing the curriculum approach shows an increasing loss for the first 120 epochs due to the curriculum, which only calculates the loss for predicted timesteps, as described in \eqref{eq:curriculum-loss}}
   \label{fig:perf-loss-plot}
   \vspace{-10pt}
\end{figure}

Figure \ref{fig:ablation-diagram} shows the qualitative difference between the two methods, where the LSTM predicts an almost straight line in a curve. In Figure \ref{fig:perf-loss-plot}, we see that although the training loss of the LSTM is very close ($\pm 1\times 10^{-4}$), the test ADE is very different, higher by $\approx 0.03$ meters, and the IoU, which takes heading error into account, is $0.12$ higher. The dynamics equations seem to be acting as a regularizer, discouraging overfitting to the training dataset through enforcement of system dynamics. PCMP trains quickly, obtaining most of its performance by epoch $150$ without curriculum, and epoch $175$ with the curriculum. This shows a direction for future work, where PCMP could be trained online with data collected in a deployment.

\subsubsection{Conformal Prediction Coverage}
We used the validation dataset as the calibration dataset and calculated $95\%$ regions. We calculated the single-step coverage on the test set or the percentage of trajectories where the prediction lies within the region, for the Rotated Rectangular and Frenet Regions. These are presented in Table \ref{tab:conformal-pred-coverage}. The single-step coverage is close to our desired $95\%$ coverage, with the $s$ and $s\land d$ coverage for the Frenet Region having $1-\delta$ coverage on the test dataset. All of the multi-step coverage guarantees hold, however, these regions are conservative, and are much larger than the single-step coverage regions.

\begin{table}[h]
    \vspace{-10pt}
    \caption{Conformal Prediction Region coverage on Test Set}
    \label{tab:conformal-pred-coverage}
    \centering
    \begin{tabular}{c|c|W{c}{1.4cm} | W{c}{1.4cm}}
       \parbox[c]{1.3cm}{\centering Region} & \parbox[c]{1.3cm}{\centering Dimension}     & \parbox[t]{1.4cm}{\centering Single-Step \\Coverage}  & \parbox[t]{1.3cm}{\centering Multi-Step \\ Coverage} \\
       \hline
       \hline
       \multirow{3}{*}{\parbox[c]{1.3cm}{\centering Rotated\\Rectangle}}   & $x$           & $94.69\%$             & $\textbf{99.61\%}$ \\
                                            \cline{2-4}
                                            & $y$           & $94.86\%$             & $\textbf{99.68\%}$ \\
                                            \cline{2-4}
                                            & $x\land y$   & $94.95\%$                     & $\textbf{99.84\%}$ \\
        \cline{1-4}
       \multirow{3}{*}{Frenet}              & $s$           & $\textbf{95.52\%}$             & $\textbf{98.65\%}$ \\
                                            \cline{2-4}
                                            & $d$           & $94.87\%$             & $\textbf{98.59\%}$ \\
                                            \cline{2-4}
                                            & $s\land d$   & $\textbf{95.55\%}$             & $\textbf{99.49\%}$ \\
       \hline
    \end{tabular}
   %performance-comparison/(LSTM-CURVATURE-8|PIMP-30.*-8-.*-CURR-1)
   \vspace{-10pt}
\end{table}

\begin{table*}[t]
    \caption{Results showing average and final displacement error for approaches trained on all race lines and evaluated on all race lines}
    \label{tab:performance-results}
    \centering
    \begin{tabular}{c|c|c|c|c}
       Approach             & Training Loss ($\downarrow$)  & ADE ($\downarrow$)                & FDE ($\downarrow$)                & IoU ($\uparrow$) \\
       \hline
       CTRV (Physics)       &  $0.006068$                   & $0.25683 \pm 0.001991$            & $0.74000 \pm 0.005637$            & $0.43644 \pm 0.002339$ \\
       LSTM                 &  $\textbf{0.001212}$          & $0.07740 \pm 0.000149$            & $0.14845  \pm 0.000171$           & $0.66444 \pm 0.000637$ \\
       PCMP                 &  $0.001501$                   & $0.04917 \pm  0.000204$           & $0.12023 \pm 0.000397$            & $0.78254 \pm 0.000521$  \\
       PCMP + Curriculum    &  $0.001451$                   & $\textbf{0.04539} \pm 0.000160$   & $\textbf{0.10644} \pm 0.000275$   & $\textbf{0.79703} \pm 0.000719$ \\
       \hline
    \end{tabular}
   %performance-comparison/(LSTM-CURVATURE-8|PIMP-30.*-8-.*-CURR-1)
   \vspace{-10pt}
\end{table*}

\subsection{Robustness to Out of Distribution Driving}
\label{sec:ood-driving}
To examine the robustness of the Physics Constrained motion prediction algorithm to out-of-distribution (OOD) behavior, we train on the centerline offset race lines but predict on the optimal race line, which has significantly different behavior through turns, as pictured in Figure \ref{fig:dataset-racelines}. We report the ADE, FDE, and IoU on the in-distribution (ID) test set and the OOD test set in Table \ref{tab:robustness-results}. No conformal prediction regions are used here, as the validation dataset does not have the same distribution as the OOD behavior.

The results show that PCMP with a curriculum approach obtains the ID and OOD errors. The performance of the LSTM gets worse for the OOD test set by a factor of $\sim 2$, but both PCMP predictors perform better on the OOD test set than the ID test set. This could be due to the optimal race line having fewer sharp turns than the center line offsets, and carrying higher velocity; both of these scenarios would advantage a model that had a notion of inertia. Examining Figure \ref{fig:lstm-ood-error} shows that the LSTM prediction includes sharp rotations in a straight, behavior that is largely aberrant. PCMP for the same scenario predicts a trajectory much closer to the true trajectory.

\begin{table*}[h]
   \caption{Results showing the robustness of the above approaches to OOD data. \\\footnotesize{OOD=Out-Of-Distribution,ID=In-Distribution}}
   \label{tab:robustness-results}
   \centering
   \begin{tabular}{c|c|c|c|c|c|c}
       Approach            & ID ADE ($\downarrow$)    & ID FDE ($\downarrow$)   & ID IoU ($\uparrow$) & OOD ADE ($\downarrow$)    & OOD FDE ($\downarrow$)    & OOD IoU ($\uparrow$) \\
       \hline
       CTRV (Physics)    &  $0.2541$ & $0.7322$ & $0.4373$              & $0.2679$ & $0.7717$ & $0.4306$ \\
       LSTM              &  $0.0698$ & $0.1411$ & $0.7027$              & $0.1528$ & $0.3026$ & $0.5312$ \\
       PCMP              &  $0.0593$ & $0.1517$ & $0.7495$              & $0.0559$ & $0.1468$ & $0.7409$ \\
       PCMP + Curriculum &  $\textbf{0.0513}$ & $\textbf{0.1235}$ & $\textbf{0.7748}$              & $\textbf{0.0466}$ & $\textbf{0.1149}$ & $\textbf{0.7696}$\\
       \hline
   \end{tabular}
   \vspace{-10pt}
   %performance-comparison/(LSTM-CURVATURE-8|PIMP-30.*-8-.*-CURR-1)
\end{table*}

\subsection{Robustness to Modeling Error}
\label{sec:model-error}
\begin{figure}[h]
   \centering
   \includegraphics[width=0.75\linewidth]{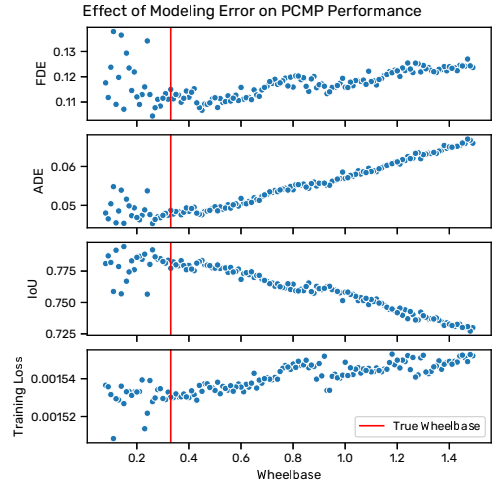}
   \vspace{-12pt}
   \caption{The plot shows the ADE, FDE, training, and testing losses of a model vs the wheelbase estimate used in the model. The true wheelbase is denoted by the red line.}
   \label{fig:modeling-error}
   \vspace{-13pt}
\end{figure}

To understand the impact of modeling errors on PCMP performance, we initialize PCMP with incorrectly estimated wheelbases. PCMP is trained using the curriculum approach where the number of epochs per input is 1. One model is trained for each wheelbase from the set $\{0.0802, 0.0902, \ldots, 1.4902, 1.5002\}$. We record the ADE, FDE, IoU, and training loss of each model. Our results are presented in Figure \ref{fig:modeling-error}. We can see that there is a linear correlation between the wheelbase and our tracked metrics. This is interesting, as we would expect the error metrics to be the lowest for the true wheelbase. We find linear correlations between all of the reported metrics and the wheelbase. The R-values of these correlations are $0.95$ for ADE, $-0.93$ for IoU, and $0.83$ for training error. The FDE error is correlated with an R-value of $0.51$.

We were unable to sample lower values for the wheelbase due to numerical instability in the training. This comes from the term $V\tan(\delta)/L$ in the dynamics. If we create two bicycle models, one with wheelbase $L$, and one with wheelbase $L/N$, where $N>1$, we can show that all states reachable by the first model can be reached by the second model with the smaller wheelbase. Intuitively, we can see that a unicycle is more maneuverable than a bicycle, as a unicycle can turn on a dime, whereas a bicycle cannot. We must weigh numerical instability in training with prediction accuracy and model fidelity when choosing a wheelbase. More importantly, we need not have perfect wheelbase estimation, we can simply underestimate and get accurate results. This analysis is only applicable to the bicycle model, however, the bicycle model can be regarded as a representative model for vehicle motion.

\subsection{Limitations}
While PCMP shows several benefits, there are limitations, which can be placed in three categories; model selection, tolerance to large measurement noise, and conformal prediction under distribution shift.

First, in our autonomous vehicle experiments, we know that the bicycle model is a valid model for the system we predict for. In the real world, if systems with four-wheel steering are used, the same model may not produce accurate predictions. This can be remedied by collecting data to use to build prediction regions for this case or using Adaptive Conformal Prediction \cite{dixit_adaptive_2022}. It can also be addressed by using a more apt model, but this can not be done online.

Next, we have shown that our method is tolerant to relatively small amounts of noise, but under heavy noise regimes, PCMP may fail to provide state-of-the-art predictions. We used PCMP paired with LaneGCN \cite{liang_learning_2020}for feature extraction on the Argoverse Dataset \cite{chang_argoverse_2019} and achieved a minADE(K=6) of $1.86$, to vanilla LaneGCN's $0.87$. This was largely due to the high amounts of noise in the dataset, paired with the lack of heading and velocity information in the dataset, which we addressed by using LSTMs to estimate these values. This does however show the potential for PCMP to be paired with complex context extraction and intent estimation networks.

Finally, we make the assumption that we have a calibration dataset with data i.i.d to our testing dataset in order to generate valid Conformal Prediction Regions. This assumption may often not hold, especially under distribution shifts. Techniques like Adaptive Conformal Prediction \cite{dixit_adaptive_2022} can be used to relax or eliminate this assumption, where the prediction regions adapt in real time to the data observed by the system.

\section{Conclusion}
We have shown that PCMP provides guarantees of the dynamic feasibility of its' predicted trajectories in Section \ref{sec:dyn-feas-guarantee}. We have shown how control inputs are predicted and can be interpreted to understand predicted intent. We have shown that PCMP performs better than baselines in \ref{sec:performance} and that it is better able to adapt to out-of-distribution behavior in \ref{sec:ood-driving}. Section \ref{sec:model-error} shows that underestimates of the wheelbase parameter can provide similar or better performance. Two new conformal prediction regions were developed in \ref{sec:cp-regions}, for use in scenarios with and without map information.

% \section*{Acknowledgment}
\bibliographystyle{IEEEtran}
\bibliography{IEEEabrv,references}

\newpage
\appendices

\end{document}